%% file: colm2025_conference.tex
\documentclass{article} % For LaTeX2e
\usepackage[preprint]{colm2025_conference}
\usepackage{xcolor}[x11names]
\usepackage{microtype}
\usepackage{hyperref}
\usepackage{url}
\usepackage{booktabs}

\usepackage{subcaption}

\usepackage{xcolor}    % for \colorbox and color control
\usepackage{graphicx} 

\usepackage{lineno}
\usepackage[T1]{fontenc}
\usepackage{makecell}

\usepackage{times}
\usepackage{latexsym}
\usepackage{placeins}
\usepackage{longtable} % For tables that span multiple pages (optional)
\usepackage{tabularx}  % For automatic column width adjustment
\usepackage{amsmath}   % For better math formatting
\usepackage[utf8]{inputenc}
\usepackage{inconsolata}
\usepackage{amsfonts}
\usepackage{float}
\usepackage{pgfplots}
\usepackage{wrapfig}
\usepackage{enumitem}
\usepackage{multirow}

\definecolor{darkblue}{rgb}{0, 0, 0.5}
\hypersetup{colorlinks=true, citecolor=darkblue, linkcolor=darkblue, urlcolor=darkblue}

\newcommand{\dataname}{\textsc{AI-Faker}}
\newcommand{\foundationmodel}{\textsc{GenAI}}
\newcommand{\AIgeneratedimages}{\textsc{Diffusion-generated images}}
\newcommand{\AImodifiedimages}{\textsc{Face-swapped images}}
\newcommand{\AItextcompletion}{\textsc{AI-text-responding}}
\newcommand{\AIpaperreview}{\textsc{AI-paper-reviewing}}

\title{Could AI \emph{Trace} and \emph{Explain} the Origins of AI-Generated Images and Text? }

\author{
        \centerline{Hongchao Fang$^1$, Yixin Liu$^2$, Jiangshu Du$^3$, Can Qin$^4$, Ran Xu$^4$, Feng Liu$^5$
         }  \\ \centerline{\textbf{Lichao Sun$^2$, Dongwon Lee$^1$, Lifu Huang$^6$, Wenpeng Yin$^1$}}
    \\
    \centerline{
    $^1$Penn State University, 
    $^2$Lehigh University,
    $^3$University of Illinois Chicago
    } \\
    \centerline{$^4$Salesforce Research,
    $^5$Drexel University, 
    $^6$UC Davis} \\
        \centerline{\{hpf5161, wenpeng\}@psu.edu}}

\begin{document}

\ifcolmsubmission
\linenumbers
\fi

\maketitle

\begin{abstract}
AI-generated content is becoming increasingly prevalent in the real world, leading to serious ethical and societal concerns. For instance, adversaries might exploit large multimodal models (LMMs) to create images that violate ethical or legal standards, while paper reviewers may misuse large language models (LLMs) to generate reviews without genuine intellectual effort. While prior work has explored detecting AI-generated images and texts, and occasionally tracing their source models, there is a lack of a systematic and fine-grained comparative study. Important dimensions—such as AI-generated images vs. text, fully vs. partially AI-generated images, and general vs. malicious use cases—remain underexplored. Furthermore, whether AI systems like GPT-4o can explain why certain forged content is attributed to specific generative models is still an open question, with no existing benchmark addressing this. To fill this gap, we introduce \dataname, a comprehensive multimodal dataset with over 280,000 samples spanning multiple LLMs and LMMs, covering both general and malicious use cases for AI-generated images and texts. Our experiments reveal two key findings: (i) AI authorship detection depends not only on the generated output but also on the model's original training intent; and (ii) GPT-4o provides highly consistent but less specific explanations when analyzing content produced by OpenAI's own models, such as DALL-E and GPT-4o itself.\footnote{\href{https://github.com/CosimoFang/AI-FAKER}{https://github.com/CosimoFang/AI-FAKER} }
\end{abstract}

\input{sections/figures_and_tables}

\section{Introduction}
The rapid growth of \foundationmodel, including large language models (LLMs) and large multimodal models (LMMs),  has led to the widespread adoption of AI-generated content across both textual and visual modalities. Although LLMs and LMMs introduce unprecedented creative and practical capabilities—from creative expression to practical automation—these same capabilities raise pressing concerns regarding authenticity, security, and ethical usage \citep{hassanin2024comprehensiveoverviewlargelanguage}. Because of the relative ease with which realistic content can be synthesized, malicious actors are now better positioned to produce deceptive materials \citep{cui2023facetransformerhighfidelity}, which in turn complicates efforts to safeguard information integrity.

In response, significant research efforts have been devoted to developing automated methods for distinguishing human-authored content from AI-generated text \citep{solaiman2019releasestrategiessocialimpacts,mitchell2023detectgptzeroshotmachinegeneratedtext,detect}, images \citep{asnani2022proactive,sha2023defakedetectionattributionfake,ding2025fairadapter}, and multimodal data \citep{DBLPHuangDYC24,DBLP04906}. However, several key research questions remain underexplored:

\begin{itemize}[leftmargin=*, itemsep=0pt]
    \item While plenty of previous studies have detected AI-generated images, a finer-grained question arises within the image modality: \textbf{When comparing fully AI-generated images (e.g., diffused from text prompts) to partially AI-generated images (e.g., face-swapped images), which is more difficult to detect?}  

    \item Existing research primarily focuses on detection, either as a binary classification task \citep{DBLPHuangDYC24,elkhatat2023evaluating} or as AI model attribution \citep{DBLPWang0ZLM23,DBLPCavaCT24}, often supplemented with human-interpretable feature explanations. However, an open question is \textbf{whether AI models, particularly LLMs/LMMs, can explain when they attribute forged text or images to specific models}.

    \item AI-generated content is often misused for malicious purposes, such as creating fake images through face-swapping or fabricating unverifiable paper reviews. While prior work focuses on expert-designed adversarial attacks \citep{DBLPZhouH024,DBLPHuangZLYWY24}, real-world misuse typically arises naturally without deliberate attack strategies. This raises the question: \textbf{what distinct behavioral patterns emerge when comparing general and malicious use cases}?

\end{itemize}

Unfortunately, no existing dataset enables comparative research across three critical dimensions: AI-generated images vs. AI-generated text, fully AI-generated images vs. partially AI-generated images, and general use cases vs. malicious use cases. To address this gap, we introduce \dataname, a large-scale dataset designed to facilitate these fine-grained model tracing and explanation. Specifically, \dataname~supports: i) \textbf{Text \& Image Forgery Detection}: \dataname~includes AI-generated outputs from both LLMs and LMMs, creating a unified benchmark for directly comparing detectability across textual and visual modalities. ii) \textbf{Fully AI-Generated vs. Partially AI-Generated Images}: The dataset encompasses both fully diffused images from text prompts and partially modified images featuring AI-swapped faces from various LMMs. This allows for studying not only binary fake detection but also AI model tracing in today's increasingly complex digital landscape. iii) \textbf{General vs. Malicious Use Cases}: Beyond standard AI-generated content, such as text and image synthesis from prompts, \dataname~incorporates two high-stakes misuse scenarios—\emph{face swapping} and \emph{peer review generation} (evaluated on full paper submissions)—providing insights into more sophisticated and malicious real-world threats. Table \ref{tab:data_com} provides a detailed comparison between our dataset, \dataname, and other representative benchmarks for AI-generated content detection.

\tableDataset

In our experiments, we conducted extensive studies to address the proposed research questions and beyond. Our findings reveal several key insights. First, in model tracing, detecting the authorship of \AIgeneratedimages~is relatively easy, achieving around 90\% accuracy. However, identifying the source of \AImodifiedimages~is much more challenging, with performance close to random guessing. For AI-generated paper reviews (\AIpaperreview), authorship detection remains effective even when the review format is standardized, whereas detecting AI-generated text in \AItextcompletion~is notably more difficult. These comparisons suggest that AI authorship detection is influenced not only by the characteristics of the generated output but also by the model's original training objective—e.g., whether it was designed to deceive human perception. Second, regarding model explanation, we observe a broader challenge: AI models are better at identifying patterns in outputs generated by other models than in their own family, reflecting the classical difficulty of self-evaluation. This limitation has important implications for building trustworthy and self-aware AI systems.

Overall, our contributions can be summarized in three key aspects: i) We introduce \dataname, a novel benchmark designed for AI model tracing and explanation in various aspects. ii) This is the first work to conduct comparative studies in all three dimensions: AI-generated images vs. AI-generated text, fully AI-generated images vs. partially AI-generated images, and general use cases vs. malicious use cases. iii) Our findings provide valuable insights for both generative AI developers and users, strengthening digital content integrity and informing strategies to mitigate AI-generated misinformation.

\section{Related Work}

\paragraph{Image Forgery Detection.}

Recent advances in latent diffusion models (LDMs), such as DALL-E \citep{ramesh2021zeroshottexttoimagegeneration}, Midjourney\footnote{A proprietary AI image generation tool developed by the independent research lab Midjourney, Inc. \url{https://www.midjourney.com}}, and Stable Diffusion \citep{rombach2022highresolutionimagesynthesislatent}, have raised growing concerns regarding misinformation, fake news, and cybersecurity. Moreover, face-swapping models like Simswap \citep{Chen_2020} and Uniface \citep{Zhou_2023_ICCV} enable malicious actors to fabricate convincing synthetic evidence, implicating individuals in crimes or forging alibis. To mitigate the potential harms of AI-generated and AI-modified images, various detection methods have been proposed. Researchers have explored robust training frameworks based on diverse datasets \citep{bird2023cifakeimageclassificationexplainable,cozzolino2023syntheticimagedetectionhighlights,zhu2023genimagemillionscalebenchmarkdetecting,yan2024df40nextgenerationdeepfakedetection}, data augmentation strategies \citep{zhu2023genimagemillionscalebenchmarkdetecting}, patch-level detectors \citep{chen2018learningdark}, and limited receptive field techniques \citep{nataraj2019detectinggangeneratedfake}. Other efforts focused on specialized tasks such as image pair comparison to identify the forged one \citep{asnani2022proactive} or leveraging vision-language features for improved generalization to unseen generators \citep{ojha2023fakedetect}. DE-FAKE \citep{sha2023defakedetectionattributionfake} further shows that diffusion-generated images may retain subtle ``digital fingerprints," detectable through Fourier transforms \citep{leethorp2022fnetmixingtokensfourier} or attention-based vision methods, even if imperceptible to humans. Few-shot and zero-shot detection approaches have also been explored for adapting to evolving diffusion models \citep{cozzolino2024zeroshotdetectionaigeneratedimages}.

\paragraph{Text Forgery Detection.}

 Early detection studies focused on statistical cues and linguistic artifacts to differentiate machine-generated- from human-generated text. However, the emergence of advanced models such as GPT-4 \citep{openai2024gpt4technicalreport} and DeepSeek \citep{deepseekai2025deepseekr1incentivizingreasoningcapability} has challenged detector robustness, motivating the development of more comprehensive benchmarks. Recent datasets, including RAID \citep{dugan2024raidsharedbenchmarkrobust}, M4 \citep{wang2024m4multigeneratormultidomainmultilingual}, and MULTITuDE \citep{Macko_2023}, compile large-scale corpora covering multiple domains, languages, and generator outputs to systematically assess detector performance.

These benchmarks reveal that existing detectors often show poor generalization. To address this, recent methods emphasize generalizable detection techniques without assuming knowledge of the underlying LLM. These approaches focus on features such as writing style, author-specific patterns, and topology-based representations, exemplified by TopFormer \citep{zhang2022topformertokenpyramidtransformer}. Additionally, some systems offer fine-grained classification to handle partially machine-written or machine-polished content \citep{abassy2024llmdetectaivetoolfinegrainedmachinegenerated}, while others prioritize multilingual and multi-domain robustness, as showcased in recent shared tasks \citep{wang2024semeval2024task8multidomain,dugan2025genaicontentdetectiontask}.

\paragraph{Multi-modal Forgery Detection.}Several prior works have investigated the detection of multi-modal forged content \citep{abdelnabi2022opendomaincontentbasedmultimodalfactchecking, DBLPHuangDYC24}. While some focus on small-scale, human-generated multi-modal fake news \citep{10.1145/3308558.3313552}, others examine out-of-context misinformation, wherein a genuine image is paired with mismatched text—yet neither image nor text is actually manipulated \citep{abdelnabi2022opendomaincontentbasedmultimodalfactchecking}. These approaches typically address only binary classification, relying on basic image-text correlation. 

Our work differs from prior studies in three aspects:
i) We conduct an in-depth comparative study across modalities and settings, covering AI-generated images vs. text, fully vs. partially AI-generated images, and general vs. malicious use cases.
ii) We go beyond AI authorship tracing by also providing explanations for model attribution, offering insights beyond raw detection results.
iii) We highlight that such explanations are crucial for enhancing digital content protection and guiding the development of more robust generative AI models.

\section{\dataname~Construction}

\begin{figure}[t]
  \centering
  \includegraphics[width=0.9\textwidth]{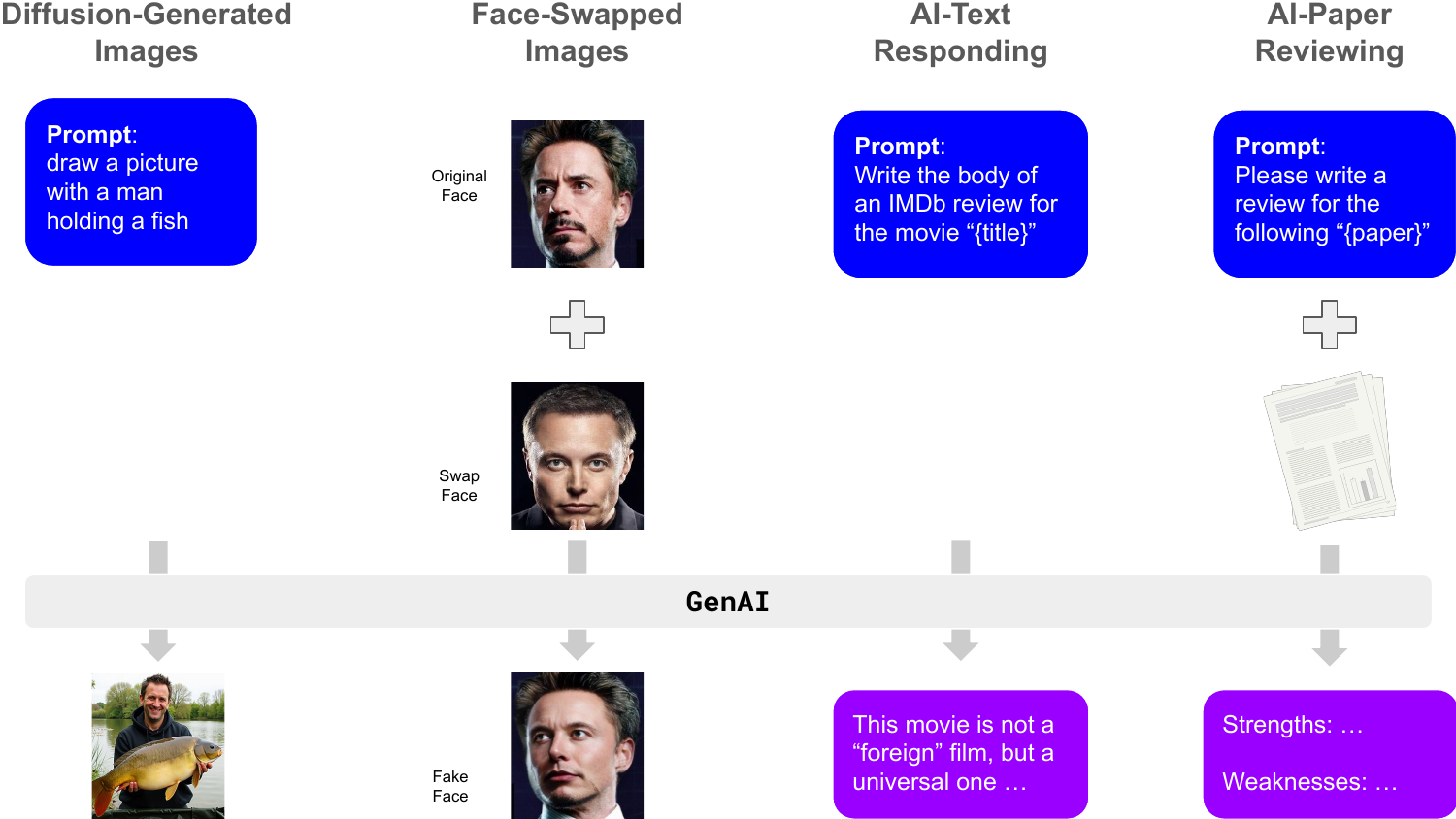}
  \caption{Illustration about the four settings in \dataname.}
\label{fig:data_source}
\end{figure}

Our \dataname~dataset covers forged content across modalities and settings (Figure \ref{fig:data_source}). For images: \AIgeneratedimages~given text prompt (\textcolor{blue}{general use case}), \AImodifiedimages~ given two images as prompt (\textcolor{red}{misuse}). For text: \AItextcompletion~given text prompt (\textcolor{blue}{general use case}), \AIpaperreview~given the full paper submission and text prompt (\textcolor{red}{misuse}).

\paragraph{\AIgeneratedimages.}
First, to construct a subset of natural images (``original''), we randomly sampled 10,000 images from ImageNet \citep{russakovsky2015imagenetlargescalevisual}. Each image is accompanied by a caption. It is worth mentioning that ImageNet images, compiled in 2015, can be safely assumed to be non-AI-generated due to the dataset's curation process and the technological limitations of generative AI at that time.

For \AIgeneratedimages, we use the caption of each natural image as input to the following five popular diffusion models to generate synthetic images: closed-source models Midjourney and DALL-E \citep{ramesh2021zeroshottexttoimagegeneration},  open-source models: Sdxl-turbo \citep{podell2023sdxlimprovinglatentdiffusion}, Stable-diffusion-xl-base-1.0 \citep{rombach2022highresolutionimagesynthesislatent}, and FLUX.1-dev \citep{chang2024fluxfastsoftwarebasedcommunication}.

\begin{figure}[t]
  \centering
  \includegraphics[width=0.9\textwidth]{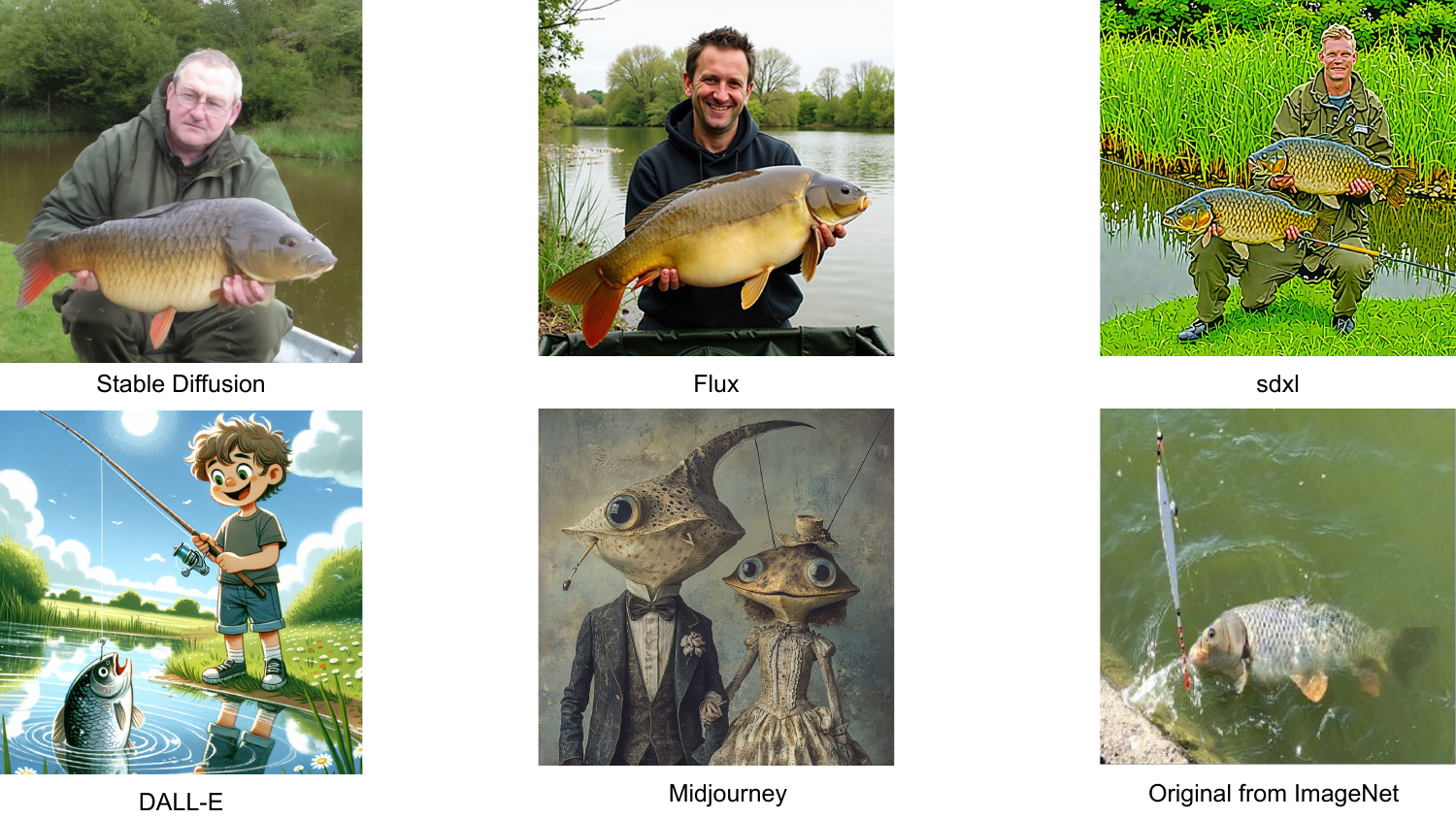}
  \caption{\AIgeneratedimages~with the prompt: \emph{draw a picture with a man fishing}.}
  \label{fig:diff}
\end{figure}

Sample prompts are provided in Appendix~\ref{app:dataconstruction}, Table~\ref{tab:diff_prompt}, and dataset statistics are summarized in Appendix~\ref{app:image}, Table~\ref{tab:AI_generated_image_dataset}. Figure~\ref{fig:diff} illustrates \AIgeneratedimages~for the prompt ``\emph{draw a picture with a man fishing},'' revealing distinct model characteristics. Notably, closed-source models often generate stylized outputs, such as paintings or cartoons, which we further analyze in later experiments.

\paragraph{\AImodifiedimages.}
We begin by collecting 6,000 images from a diverse set of sources spanning multiple domains. These images originate from old movies, TikTok, and YouTube videos created by humans, ensuring that the original class contains no AI-generated or AI-modified images. Subsequently, we apply a face detection model, YOLO5Face \citep{qi2022yolo5facereinventingfacedetector}, to filter the images, retaining only those containing faces for use in face-swapping models. The face detection model operates on images with a resolution of 640×640 pixels and employs a detection threshold of 50\%.

For face-swapping, we employ four models—Inswapper\footnote{\url{https://github.com/haofanwang/inswapper}}, SimSwap \citep{Chen_2020}, UniFace \citep{Zhou_2023_ICCV}, and BlendSwap \citep{DBLPShioharaYT23}—to modify detected 640×640 face regions. We further filter out unrecognized or blurry outputs to ensure dataset quality. The final dataset contains 28,551 images, including originals. Full details are provided in Appendix~\ref{app:image}, Table~\ref{tab:AI_modified_image_dataset}.

Figure \ref{fig:swap} in Appendix \ref{app:modify} showcases sample images from the \AImodifiedimages~subset. In these examples, the forged images often exhibit a blurry or indistinct facial area, making it difficult for humans to identify the modifying model. This observation will be further analyzed in subsequent experiments.

\paragraph{\AItextcompletion.}
To study the patterns introduced by different LLMs, we reuse a subset of the RAID dataset \citep{dugan2024raidsharedbenchmarkrobust}, which is the largest and most comprehensive benchmark for AI-generated text detection. RAID contains over 6 million samples generated by 11 different LLMs across 8 domains, 11 adversarial attack strategies, and 4 decoding techniques.
For our study, we selected 10,000 samples from each of five LLMs---Cohere, GPT-4, LLaMA, Ministral, and MPT \citep{MosaicML2023Introducing}---as well as 10,000 human-authored samples. These were drawn from all eight domains included in RAID.

\paragraph{\AIpaperreview.}
For human-generated reviews, we utilize 10,000 collected papers from OpenReview, each accompanied by expert-verified human reviews, as released by \cite{DBLPDuW0D0LZVZSZGL024}.  

For LLM-generated reviews, we prompt mainstream LLMs to generate reviews for these papers. Our approach includes both closed-source and open-source LLMs to ensure diversity. The closed-source models used include GPT-4o \citep{openai2024gpt4technicalreport}, Claude-3.5-Sonnet \citep{claude}, and Gemini-1.5-Pro \citep{geminiteam2024gemini15unlockingmultimodal}, while the open-source models consist of LLaMA-3.1-8B \citep{grattafiori2024llama3herdmodels} and Ministral-8B (new state-of-the-art mistral model \citep{jiang2023mistral7b}). Additionally, we include DeepSeek-R1 \citep{deepseekai2025deepseekr1incentivizingreasoningcapability}, one of the most advanced open-source models, to further enhance dataset diversity. Each model generates 10,000 reviews, resulting in a dataset totaling 70,000 samples.  

To promote diversity and mitigate redundancy, we instruct those LLMs to generate structured reviews that systematically address key aspects, including a paper's summary, strengths, and weaknesses. Each model is also provided with a human-written example review as a reference to improve output quality. To remove format-related artifacts in classification, we use GPT-4o to reformat all reviews into single-paragraph texts without structural markers. Details of the prompts used are provided in Appendix~\ref{app:dataconstruction}, Table~\ref{tab:review_prompt}.

\section{Experiments}

The dataset of each setting is split into the $train/dev/test$ by 8:1:1. F1 score is the official metric.

\subsection{$\mathcal{Q}_1$: How do the challenges of AI authorship attribution differ between AI-generated images and AI-generated text in general?}

To answer $\mathcal{Q}_1$, we first report classification results on the four cases: \AIgeneratedimages, \AImodifiedimages, \AItextcompletion, and \AIpaperreview. Since pursing state-of-the-art is not the focus of this work, we directly report existing representative classifiers in literature. 

For image authorship detection, we employ the Vision Transformer (\textbf{ViT-L/32}) \citep{dosovitskiy2021imageworth16x16words}, \textbf{ResNet50} \citep{he2015deepresiduallearningimage}, \textbf{PNASNet} \citep{liu2018progressiveneuralarchitecturesearch}), similar as GenImage \citep{zhu2023genimagemillionscalebenchmarkdetecting}. For text, we report four classifiers employed by RAID \citep{dugan2024raidsharedbenchmarkrobust} and M4 \citep{wang2024m4multigeneratormultidomainmultilingual}: \textbf{BERT-large-uncased} \citep{devlin2019bertpretrainingdeepbidirectional}, \textbf{GPT2} \citep{solaiman2019releasestrategiessocialimpacts}, \textbf{Logistic Regression with Bert Features}, and \textbf{Fully-connect network with Bert Features}.

\begin{table}[t]
\centering
\begin{minipage}{1.0\textwidth}
    \centering

\begin{tabular}{lccccccc}
\hline
Models & Original &  Midjourney & DALL-E 3 & Stable Diffusion & sdxl & Flux & Overall\\
\hline
ViT &  0.95 & 0.89 & 0.88 & 0.78 & 0.98 & 0.83 & 0.89 \\

Resnet50 &  0.98 & 0.91 & 0.98 & 0.97 & 0.99 & 0.98 & 0.97
\\

PNASNet &  0.86 & 0.85 & 0.84 & 0.82 & 0.86 & 0.81 & 0.84
\\
\hline
\end{tabular}
    \caption{Performance of \AIgeneratedimages~(\textcolor{blue}{general use}).}
    \label{tab:result:AIgeneratedimages}
\end{minipage}

\vspace{2mm}

\begin{minipage}{1.0\textwidth}
    \centering

\begin{tabular}{lccccccc}
\hline
Models & Original & inswapper128 & simswap256 &  uniface256 &  blendswap256 & Overall \\
\hline
ViT & 0.88 & 0.22 & 0.23 & 0.22 & 0.25 & 0.36\\
Resnet50  & 0.68 & 0.24 & 0.24 & 0.23 & 0.26 & 0.34\\
PNASNet & 0.60 & 0.21 & 0.21 & 0.22 & 0.23 & 0.30 
\\
\hline
\end{tabular}
    \caption{Performance of \AImodifiedimages~(\textcolor{red}{malicious use}).}
    \label{tab:result:AImodifiedimages}
\end{minipage}

\vspace{2mm}

\begin{minipage}{1.0\textwidth}
    \centering

\begin{tabular}{lccccccc}
\hline
Models & Human & Cohere & GPT4 & Llama & Ministral & MPT & overall
 \\
\hline
LR &  0.67 & 0.35 & 0.49 & 0.45 & 0.30 & 0.48 & 0.45 \\
FCN & 0.68 & 0.39 & 0.59 & 0.47 & 0.33 & 0.40 & 0.48\\
GPT2 &  0.87 & 0.37 & 0.63 & 0.72 & 0.46 & 0.49 & 0.59\\
Bert &  0.56 & 0.52 & 0.56 & 0.52 & 0.52 & 0.55 & 0.54\\
\hline
\end{tabular}
    \caption{Performance of \AItextcompletion~(\textcolor{blue}{general use}).}
    \label{tab:result:AItextcompletion}
\end{minipage}

\vspace{2mm}

\begin{minipage}{1.0\textwidth}
    \centering

\begin{tabular}{lcccccccc}
\hline
Models & Human & GPT & Claude  & Gemini & DeepSeek & Llama3 & Ministral & overall\\
\hline
LR & 0.89 & 0.74 & 0.89 & 0.82 & 0.87 & 0.93 & 0.69 & 0.83\\
FCN & 0.93 & 0.64 & 0.83 & 0.76 & 0.89 & 0.92 & 0.79 & 0.83 \\
GPT2 & 0.77 & 0.60 & 0.73 & 0.63 & 0.56 & 0.73 & 0.28 & 0.62\\
Bert & 0.78 & 0.40 & 0.74 & 0.59 & 0.42 & 0.79 & 0.33 & 0.58\\
\hline
\end{tabular}
\end{minipage}
    \caption{Performance of \AIpaperreview~(\textcolor{red}{malicious use}).}
    \label{tab:result:unifiedreview}
\vspace{-20pt}
\end{table}

Here, we present four tables together for analysis: Table \ref{tab:result:AIgeneratedimages} (\AIgeneratedimages), Table \ref{tab:result:AImodifiedimages} (\AImodifiedimages), Table \ref{tab:result:AItextcompletion} (\AItextcompletion), and Table \ref{tab:result:unifiedreview} (\AIpaperreview~with unified-format). From these results, we derive the following observations.

Regarding general versus malicious use cases, image and text classification exhibit entirely different behaviors. Detecting the authorship of \AIgeneratedimages~is relatively straightforward, with performance reaching approximately 90\%. However, identifying authorship in \AImodifiedimages~is significantly more challenging, with performance close to random guessing. In contrast, detecting AI authorship in \AIpaperreview~is relatively easy, regardless of whether the review format has been unified through rewriting. On the other hand, identifying AI-generated text in \AItextcompletion~proves to be quite difficult.

We attribute these differences to various factors. For instance, diffusion models, as illustrated in Figure \ref{fig:diff}, often exhibit superficial patterns, such as DALL-E's cartoon-like style or SDXL’s lower resolution. In contrast, \AImodifiedimages~contain only minor alterations, as shown in Appendix \ref{app:modify} Figure \ref{fig:swap}, making authorship detection challenging, particularly for human observers. Similarly, in \AItextcompletion, the responses generated by LLMs rely on pre-trained knowledge and are typically short, limiting linguistic pattern variations. However, in \AIpaperreview, the reviews are significantly longer, allowing more room for detectable superficial patterns to emerge.

% Furthermore, it is essential to consider the intended purpose of these models. Face-swapping models and text-completing LLMs are specifically designed to deceive human perception—face-swapped images aim to appear indistinguishable from real ones, and LLM-generated responses are optimized to mimic human-like text. Consequently, detecting their authorship is inherently difficult. In contrast, general diffusion models and standard LLMs are not explicitly trained to deceive. Diffusion models focus on visualizing textual prompts, while general LLMs generate human-like text without necessarily adhering to strict review criteria. Thus, \textbf{AI authorship detection depends not only on the characteristics of the generated output but also on the original intent behind the AI model's training.}

Furthermore, it is essential to consider the intended purpose of these models. Face-swapping models and text-completing LLMs are specifically designed to deceive human perception—face-swapped images aim to appear indistinguishable from real ones, and LLM-generated responses are optimized to mimic human-like text. Consequently, detecting their authorship is inherently difficult. In contrast, general diffusion models and standard LLMs are not explicitly trained to deceive. Diffusion models focus on visualizing textual prompts, while general LLMs generate human-like text without strictly adhering to review criteria. Thus, \textbf{AI authorship detection depends not only on the characteristics of the generated output but also on the original intent behind the AI model's training.}

\subsection{$\mathcal{Q}_2$: How effectively can LMMs Like GPT-4o explain AI authorship?}

To address $\mathcal{Q}_2$, we conduct experiments on both \AIgeneratedimages~and \AIpaperreview. We assume that each setting involves $N$ GenAI models. From each model, we randomly sample 5 generated instances (either images or texts). We then prompt GPT-4o, one of the strongest AI models that can handle both text and image modalities, with the following instruction (using images as an example):

\begin{quote}
\colorbox{gray!20}{\parbox{\linewidth}{%
We now know that  images [$image_1^1$, $image_2^1$, $\cdots$,  $image_5^1$] are generated by $LMM_1$,  images [$image_1^2$, $image_2^2$, $\cdots$,  $image_5^2$] are generated by $LMM_2$, $\cdots$,  images [$image_1^N$, $image_2^N$, $\cdots$,  $image_5^N$] are generated by $LMM_N$. Please compare the images across those LMMs and identify distinguishing features that explain why specific images are attributed to their respective models. }}
\end{quote}

\begin{table}[t]
\centering
\begin{tabular}{>{\centering\arraybackslash}m{1cm}|p{5cm}|p{5.5cm}}
\hline
\textbf{Batch} & \multicolumn{1}{c|}{\textbf{DALL-E3}} & \multicolumn{1}{c}{\textbf{Stable Diffusion}} \\
\hline
\multirow{4}{*}{1} & DALL-E tends toward whimsical or heightened-expression scenes and
slightly simplified backgrounds. & Stable Diffusion merges a decent level of realism with softer, diffused textures and repeating pattern artifacts (the floral backdrop in
the woman’s portrait) \\
\hline
\multirow{3}{*}{2} & DALL-E Cheerful, cartoon-like image with big eyes, bright colors, and
soft transitions & Stable Diffusion An anime-style portrait with a subdued palette, softly
diffused brushstrokes, and stylized proportions. \\
\hline
\end{tabular}
\caption{Examples of GPT4o's explanation to forged images by DALL-E and Stable Diffusion.}
\label{tab:explanexample}
\end{table}

We repeat this process for $B$ batches, where each batch consists of randomly sampled AI-generated data. This results in a total of $N \times B$ explanation texts, one for each instance batch generated by the corresponding LLM/LMM. The details of the explanations can be found in Appendix \ref{app:explain}, Table \ref{tab:image_desc} and Table \ref{tab:text_desc}, and an example snippet in Table \ref{tab:explanexample}. To evaluate the quality of these explanations, we introduce two quantitative metrics:

\begin{itemize}[leftmargin=*, itemsep=0pt]
    \item \textbf{Specificity}: Measuring whether GPT-4o's explanation for a given LMM/LLM is specific to that model (i.e., it is undesirable if the same explanation also frequently applies to other models). 
    
    We compute \emph{Specificity} for each of the $B$ explanations of a given GenAI model. Inspired by the ``IDF'' component in ``TF-IDF'', for each explanation, we compute its cosine similarity (using SentenceBERT~\citep{DBLPReimersG19}) with the $B$ explanations of every other GenAI model and record the maximum similarity. We then sum these maximum similarities and take the inverse to obtain the specificity score. This process yields $B$ specificity scores for each GenAI model. \emph{Higher specificity indicates better trustworthiness}.

    \item \textbf{Variation}: Evaluating how consistent GPT-4o's explanations are across the $B$ batches of randomly sampled AI-generated data (i.e., large variation is undesirable since the explanation should remain stable when sampling different data from the same model).
    
    To compute \emph{Variation}, we first calculate the pairwise cosine similarities among the $B$ explanations of each GenAI model. These similarities are then converted to distances by taking $1.0 - \text{similarity}$. This results in $B \times (B-1)$ variation scores for each GenAI model. \emph{Lower variation indicates better trustworthiness}.
\end{itemize}

\begin{figure}[t]
    \centering
    \begin{subfigure}[t]{0.45\textwidth}
        \centering
        \includegraphics[width=\linewidth]{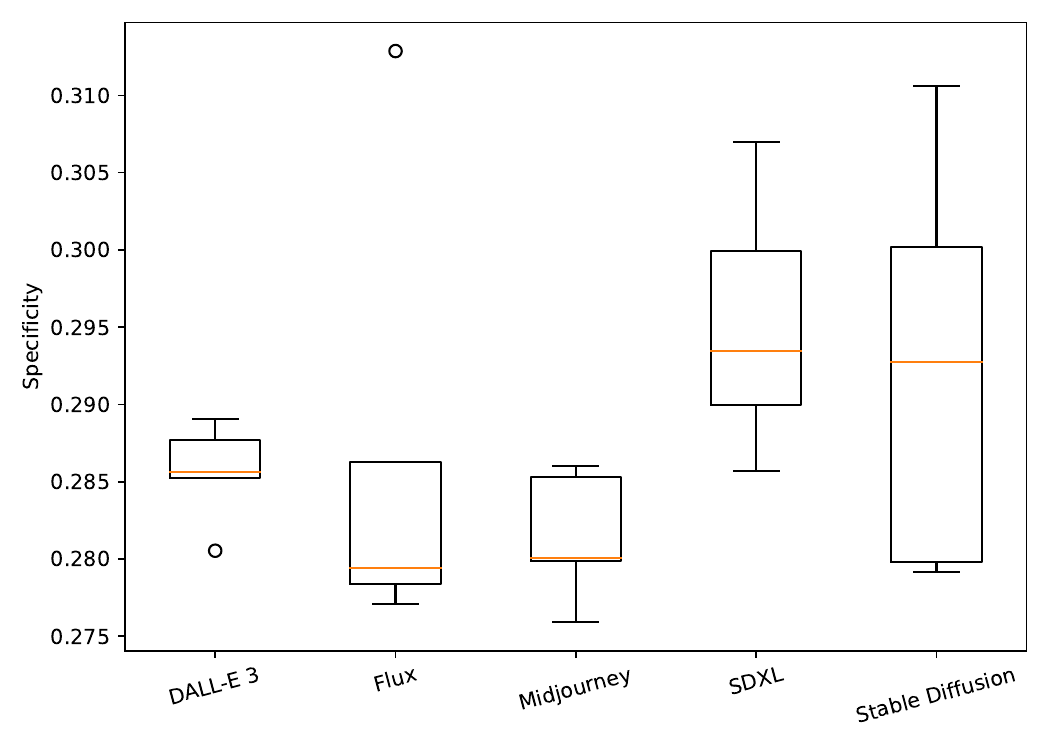}
        \caption{Specificity of explanations to LMMs}
    \end{subfigure}%
    \hspace{1cm}
    \begin{subfigure}[t]{0.45\textwidth}
        \centering
        \includegraphics[width=\linewidth]{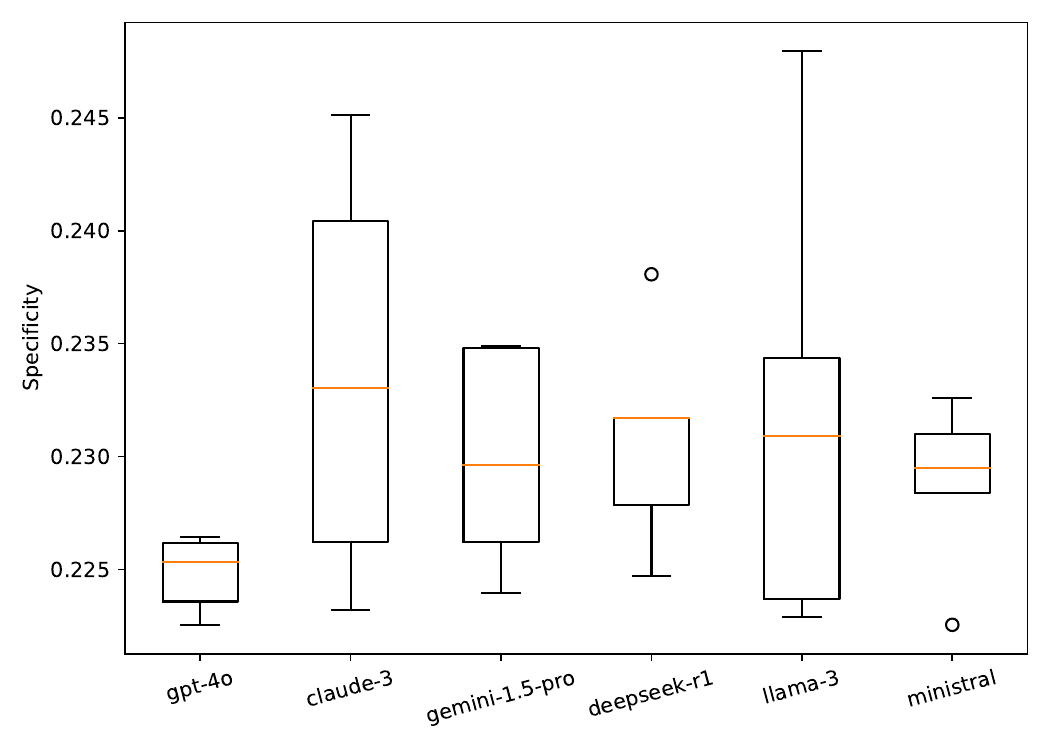}
        \caption{Specificity of explanations to LLMs}
    \end{subfigure}

    \begin{subfigure}[t]{0.45\textwidth}
        \centering
        \includegraphics[width=\linewidth]{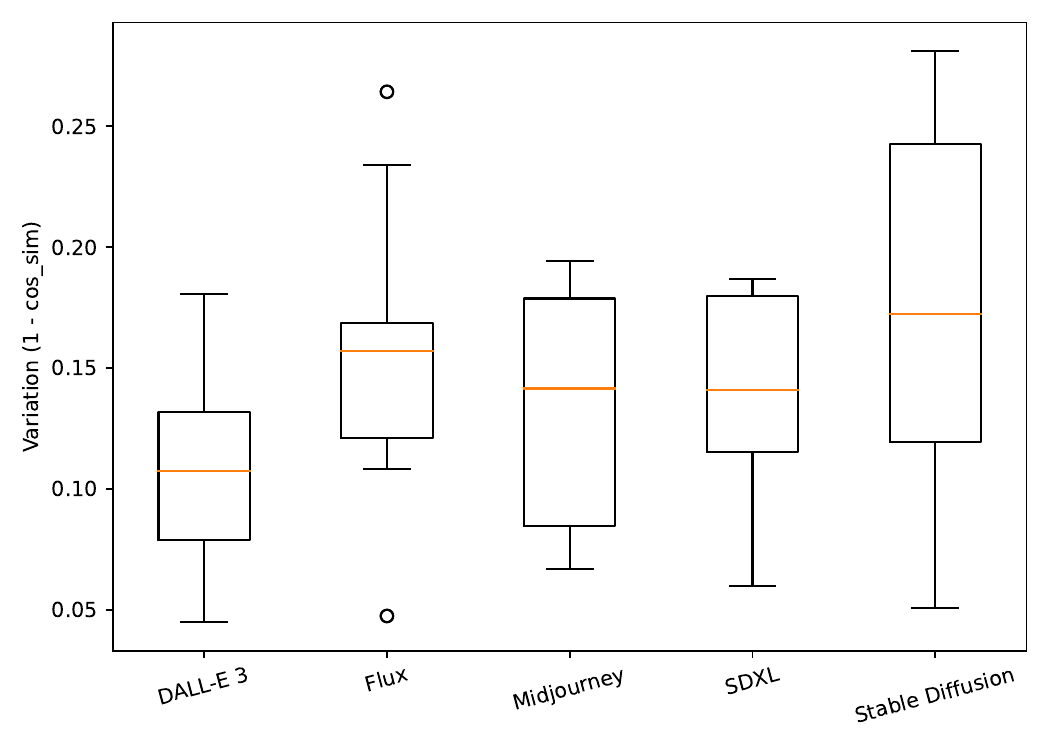}
        \caption{Variations of explanations to LMMs}
    \end{subfigure}%
    \hspace{1cm}
    \begin{subfigure}[t]{0.45\textwidth}
        \centering
        \includegraphics[width=\linewidth]{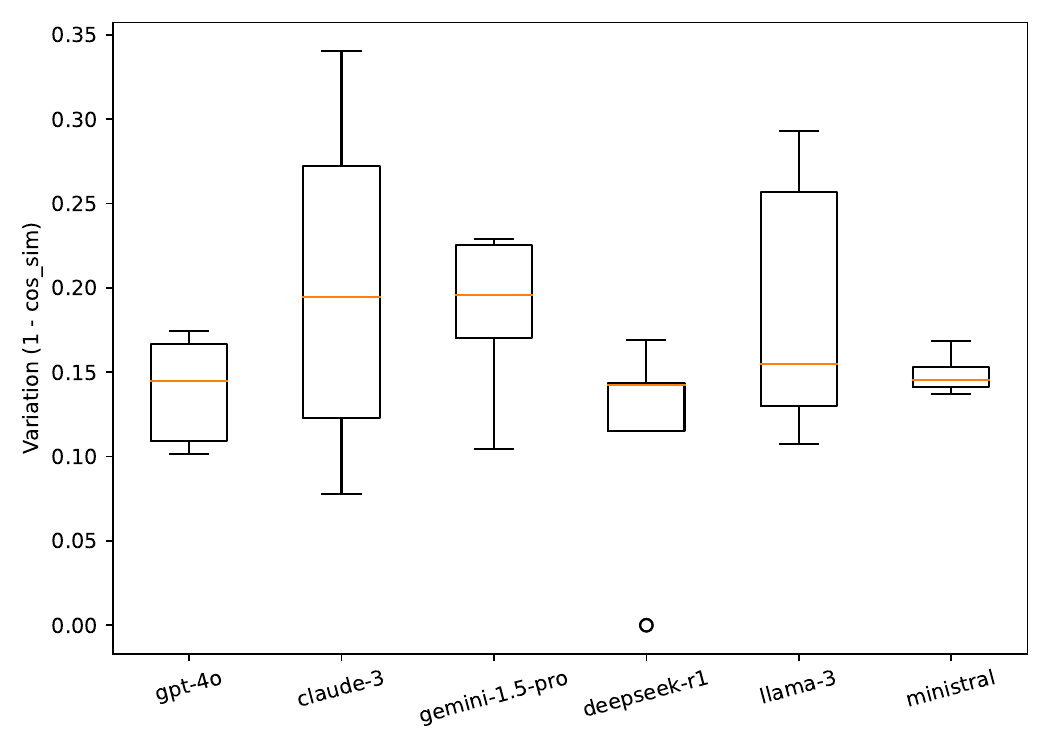}
        \caption{Variations of explanations to LLMs}
    \end{subfigure}
    \caption{Quality of GPT4o's explanations to the origins of \AIgeneratedimages~and \AIpaperreview.}
    % \captionsetup{aboveskip=2pt, belowskip=2pt}
    \label{fig:explain}
\end{figure}

 Figure~\ref{fig:explain} presents the distribution of \emph{Specificity} and \emph{Variation} for all LMMs and LLMs in both the \AIgeneratedimages~and \AIpaperreview~tasks. 
We observe the following:  
(i) For diffusion models generating images, SDXL and Stable Diffusion exhibit the highest average specificity, suggesting that their generated features are more distinctive and easier for GPT-4o to detect. In contrast, GPT-4o provides the most consistent explanations when evaluating DALL-E outputs.  
(ii) For LLMs generating paper reviews, GPT-4o, when serving as both the generator and the judge, yields the lowest specificity, indicating that it struggles to identify features specific to its own outputs. On the other hand, GPT-4o and DeepSeek demonstrate the highest consistency across their explanations.

Interestingly, we observe a consistent pattern \textbf{when GPT-4o judges outputs from OpenAI's own models—DALL-E and GPT-4o itself—characterized by high consistency but low specificity in its explanations}. This may reflect a shared design philosophy within OpenAI models, which prioritize general, human-aligned outputs over model-distinctive features. As a result, GPT-4o produces stable explanations but struggles to identify cues unique to its own or DALL-E's generations. This effect is especially evident in self-evaluation, where GPT-4o shows difficulty distinguishing its own outputs, likely due to distributional familiarity or alignment constraints. Overall, the results point to a broader challenge: \textbf{AI models are capable of detecting patterns in outputs from others more easily than in their own, mirroring the well-known difficulty of self-evaluation}, which may have significant implications for developing trustworthy and self-aware AI systems.

\subsection{$\mathcal{Q}_3$: Any specific observations for the novel task \AIpaperreview?}

To the best of our knowledge, our \AIpaperreview~subset is the first dataset to compare human and AI-generated paper reviews for AI authorship tracing while closely mimicking the real-world peer review process. Unlike previous AI-generated review datasets, which typically use simplified inputs such as only the title and abstract \citep{dugan2024raidsharedbenchmarkrobust}, our dataset is based on full paper submissions (original submissions rather than camera-ready versions). To conduct an in-depth analysis of AI authorship detection in \AIpaperreview, we examine factors such as format and length.

Appendix \ref{app:rawreview}, Tables \ref{tab:result:rawreview} and Table \ref{tab:result:unifiedreview} compare detection performance when using AI-generated reviews under the ``format-diverse'' and ``format-unified'' settings. The results clearly show that unifying the review format makes authorship detection more challenging, with accuracy dropping from an average of over 98\% to around 70\%. This suggests that using a model like GPT-4o to rewrite reviews can effectively remove some artifacts. However, the rewritten reviews may still retain certain implicit patterns, either in formatting or content, revealing the extent to which LLMs comprehend the papers they summarize.

\begin{wrapfigure}[11]{r}{0.45\textwidth}
  \begin{center}
  \vspace{-0.2in}
    \includegraphics[width=0.44\textwidth]{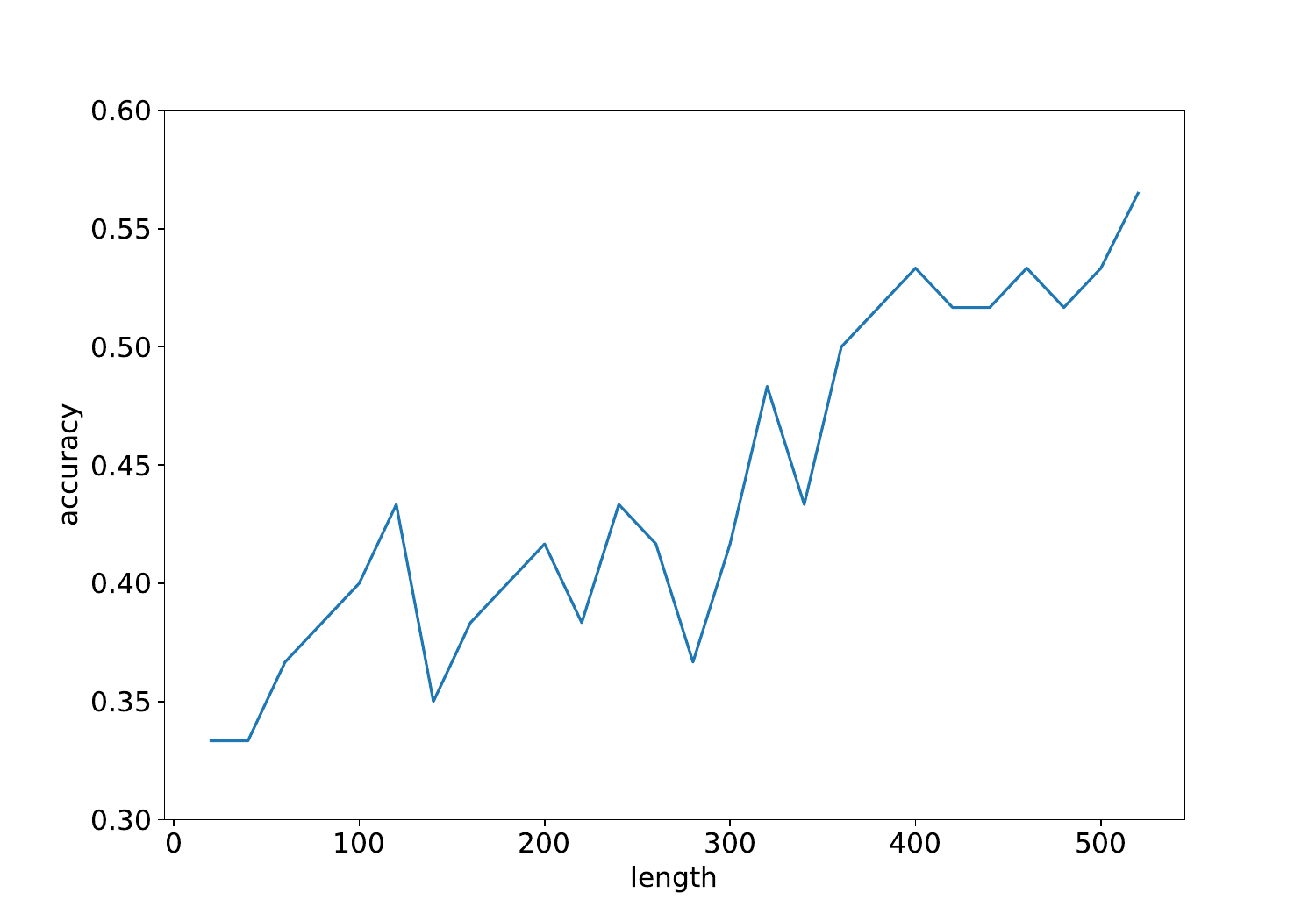}
  \end{center}
  \vspace{-5mm}
  \caption{Length effects on \AIpaperreview. }
  \label{fig:length}
\end{wrapfigure}
Additionally, since the analysis in $\mathcal{Q}_1$ hinted the impact of review length, we further investigate how length influences AI authorship detection performance. As shown in Figure \ref{fig:length}, longer reviews tend to exhibit more distinguishable features, making AI authorship easier to detect. However, the accuracy fluctuates, suggesting that other factors—such as writing style consistency and contextual coherence—may also play a role in classification performance. This conclusion endorses our analysis of \AItextcompletion~and \AIpaperreview~when responding to $\mathcal{Q}_1$.

\subsection{$\mathcal{Q}_4$: Can we discover model similarity based on LLM outputs misclassification?}

\begin{figure}[t]
    \centering
    \begin{subfigure}[t]{0.45\textwidth}
        \centering
        \includegraphics[width=\linewidth]{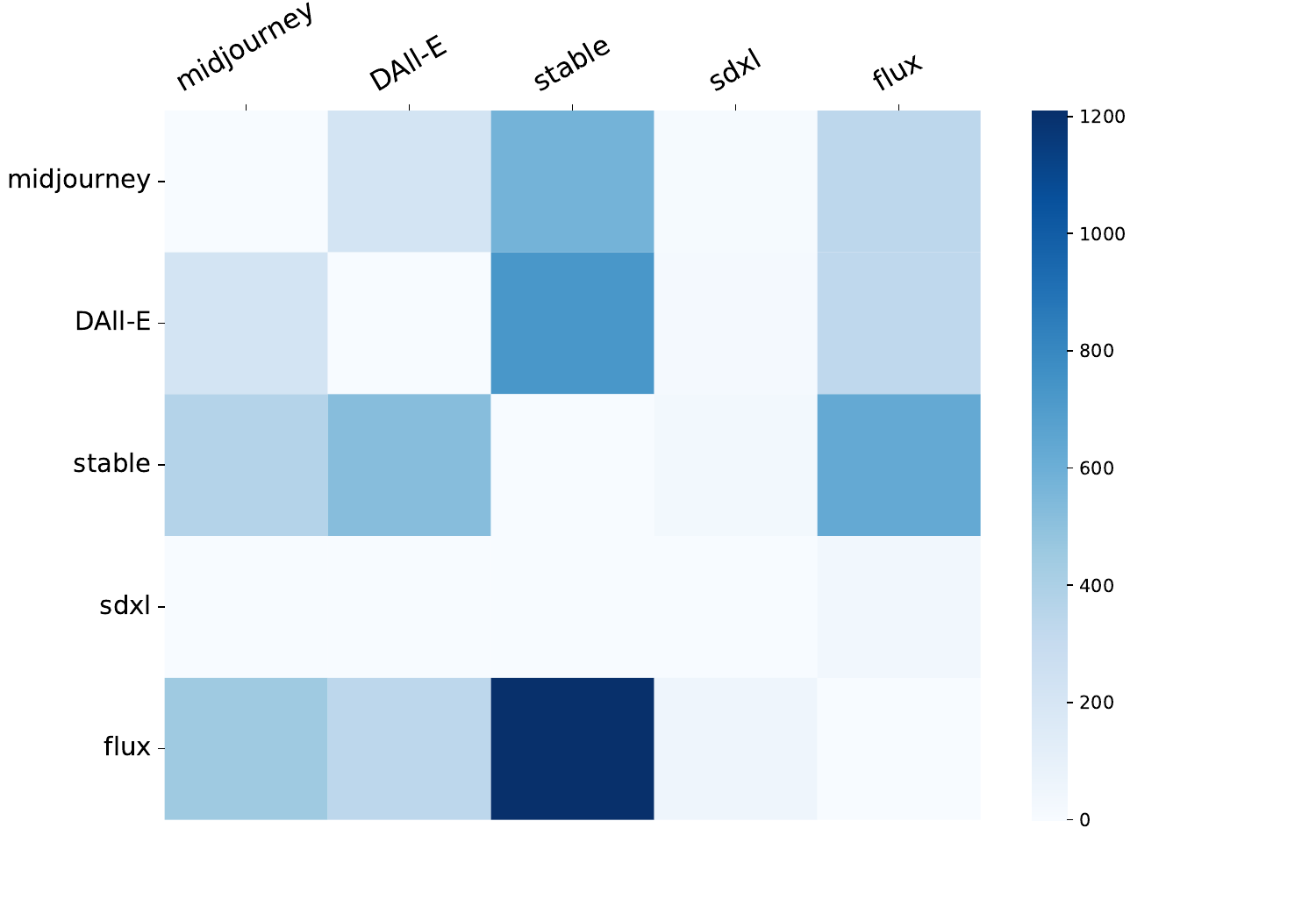}
        % \caption{Confusion matrix for tracing \AIgeneratedimages}
    \end{subfigure}%
    \hspace{1cm}
    \begin{subfigure}[t]{0.45\textwidth}
        \centering
        \includegraphics[width=\linewidth]{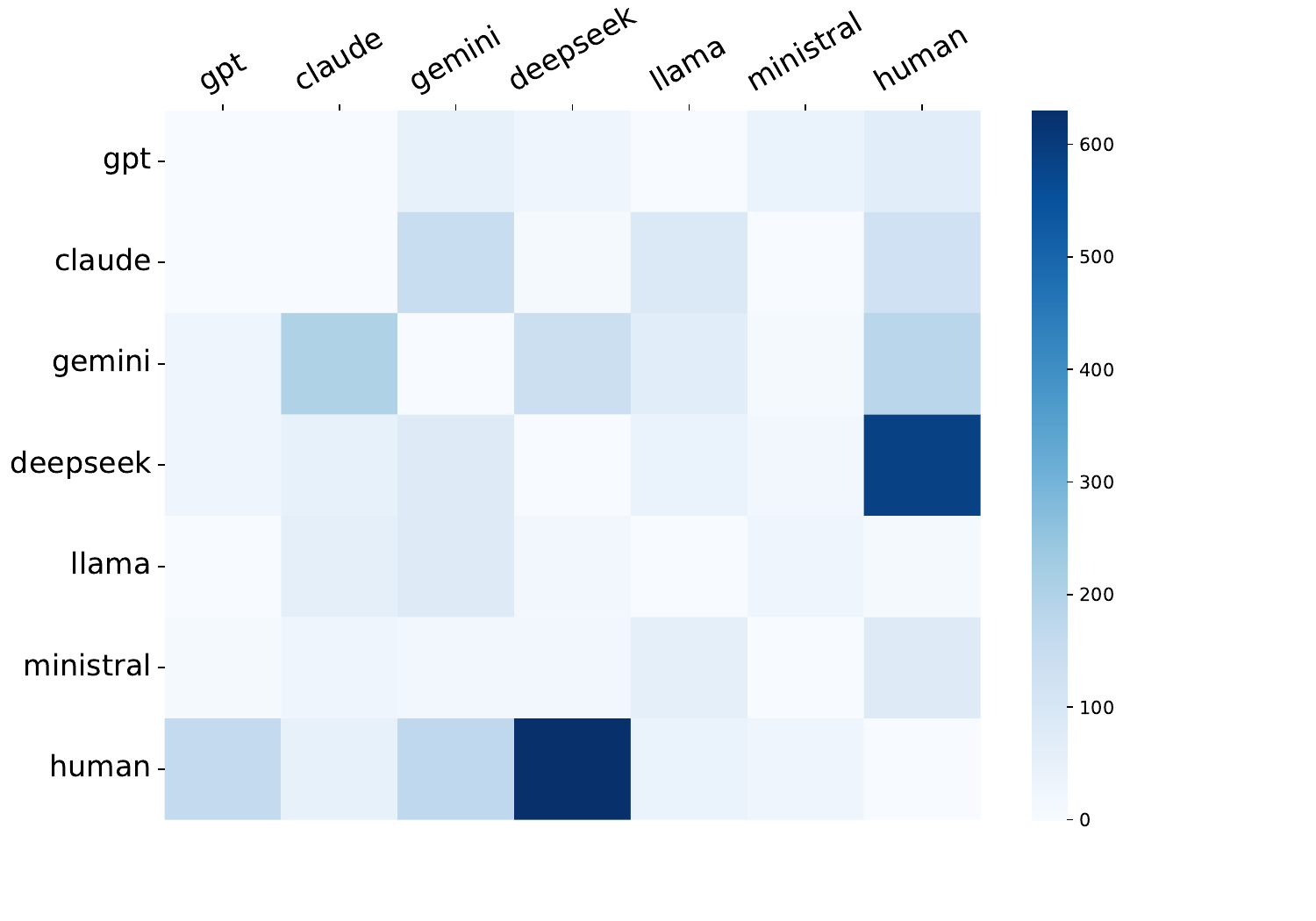}
        % \caption{confusion matrix for \AIpaperreview}
    \end{subfigure}
    \vspace{-10pt}
    \caption{Confusion matrices for tracing \AIgeneratedimages~(left) and \AIpaperreview~(right). Note diagonal values are set to 0 to highlight inter-model misclassification.}
        \vspace{-10pt}
        \label{fig:confusion}
\end{figure}

Figure \ref{fig:confusion} shows the confusion matrix when the classifier misclassifies a \AIgeneratedimages~ or \AIpaperreview. We intentionally set diagonal values to 0.0 to highlight inter-model misclassification.

From Figure \ref{fig:confusion} (left): i) \textbf{Stable Diffusion as a generalist.} We observe that images from multiple models, including MidJourney, DALL-E, and Flux, are frequently misclassified as Stable Diffusion. This suggests that Stable Diffusion acts as a dominant attractor in the classifier's decision space, likely due to its broad and general-purpose generation style. Its outputs may resemble a common visual baseline, causing the classifier to default to it when uncertain. ii) \textbf{Flux heavily overlaps with Stable Diffusion.}; iii)
\textbf{SDXL produces highly distinctive outputs.} Both the row and column corresponding to SDXL show minimal confusion, indicating that SDXL-generated images are rarely misclassified and other models are also seldom confused with SDXL. This suggests that SDXL generations possess distinctive characteristics, possibly due to unique rendering styles, textures, or fine-tuning objectives.

Figure~\ref{fig:confusion} (right) reveals that reviews generated by DeepSeek are frequently misclassified as human-written, suggesting that stronger alignment to human-like writing may blur model-specific signals. This pattern raises a concern: as LLMs become increasingly human-aligned, they may become harder to attribute. In contrast, GPT-4o and Llama exhibit more distinctive patterns, possibly due to differences in alignment strategies or generation style. These findings highlight a potential limitation of AI attribution methods, which may become less effective as models approach human-level fluency.

\section{Conclusion}

In this work, we introduce \dataname, a novel dataset designed to facilitate AI authorship detection for both AI-generated images and text. \dataname~enables a comparative study across three key dimensions: AI-generated images versus AI-generated text, fully AI-generated images versus partially AI-generated images, and general use cases versus malicious use cases. Our findings offer insights into both AI model tracing as well as explanation. We aim to help the research community better understand LLM/LMM generation behaviors and the challenges of maintaining digital integrity.

% \section*{Limitations}

% While our study shows promising results in tracing AI-generated content to its source model, it has several limitations. Our classification relies on general models like ViT for images and BERT for text, which, while effective, are not specialized for AI-generated data detection. Dedicated classification models could enhance performance. Additionally, for general AI-generated text detection, we used the RAID dataset, which was not initially designed for multi-model classification. Future work will involve updating the dataset with more diverse AI-generated texts tailored for this task.

\bibliography{colm2025_conference}
\bibliographystyle{colm2025_conference}

\appendix
\section{Appendix}

\input{sections/appendix}

\end{document}

%% file: sections/figures_and_tables.tex
\newcommand{\tableDataset}{
\begin{table}[h]
\centering
\small
\begin{tabular}{l|c|c|c|c|c|c}
\hline
Models & Size & \makecell{Domain \\diversity }& \makecell{LLM/LMM\\ coverage} &  \makecell{AI-gen\\ images} & \makecell{AI-gen\\ text}& \makecell{Natural\\ misuse}\\
\hline
M4 \citep{wang2024m4multigeneratormultidomainmultilingual} & 122k & \Checkmark & \Checkmark &    & \Checkmark & \\ 
MULTITuDE \citep{Macko_2023} & 74k &   & \Checkmark &   & \Checkmark& \\
RAID \citep{dugan2024raidsharedbenchmarkrobust} & 6.2M & \Checkmark & \Checkmark  &  & \Checkmark& \\
AuText2023 \citep{sarvazyan2023overviewautextificationiberlef2023} & 160k & \Checkmark &   &    & \Checkmark& \\
DE-FAKE \citep{sha2023defakedetectionattributionfake} & 20k &\Checkmark & \Checkmark & \Checkmark &   & \Checkmark \\
GenImage \citep{zhu2023genimagemillionscalebenchmarkdetecting} & 1.3M &\Checkmark & \Checkmark &  \Checkmark &   & \Checkmark \\
MiRAGeNews \citep{huang-etal-2024-miragenews} & 12k & &  & \Checkmark & \Checkmark & \Checkmark \\ 
\hline
\dataname & 287k &\Checkmark & \Checkmark  & \Checkmark & \Checkmark & \Checkmark\\
\hline
\end{tabular}
\caption{\dataname~vs. other representative datasets for detection of AI-generated content.}
\label{tab:data_com}
\end{table}
}

\newcommand{\tableDiffResult}{
\begin{table*}[h]
\centering
\begin{tabular}{lccccccc}
\hline
Models & Original &  Midjourney & DALL-E 3 & Stable Diffusion & sdxl & Flux & Overall\\
\hline
F1 &  95.8 & 89.1 & 88.1 & 78.3 & 98.9 & 83.6 & 89.0 \\

recall &  96.3 & 89.6 & 89.0 & 74.4 & 98.8 & 86.6 & 89.12
\\

precision &  95.3 & 88.6 & 87.3 & 82.7 & 99.1 & 80.8 & 88.97
\\
\hline
\end{tabular}
\caption{AI-generated image multi-classification results(reported with F1 Score)}
\label{tab:diff_result}
\end{table*}
}

\newcommand{\tableModifyResult}{
\begin{table*}[h]
\centering
\begin{tabular}{lccccccc}
\hline
Models & Original & inswapper128 & simswap256 &  uniface256 &  blendswap256 & Overall \\
\hline
F1 & 0.88 & 0.22 & 0.23 & 0.23 & 0.25 & 0.36\\
recall  & 0.92 & 0.24 & 0.24 & 0.23 & 0.26 & 0.38\\
precision & 0.84 & 0.21 & 0.21 & 0.22 & 0.23 & 0.34 
\\
\hline
\end{tabular}
\caption{AI-modified image multi-classification results(reported with F1 Score)}
\label{tab:edit_result}
\end{table*}
}

\newcommand{\tableReviewSelf}{
\begin{table*}[h]
\centering
\small
\begin{tabular}{lccccccc}
\hline
Models &  GPT & Claude  & Gemini & DeepSeek & Llama3 & minitral & overall\\
\hline
accuracy & 0.51 & 0.74 & 0.89 & 0.82 & 0.87 & 0.93 & 0.69 \\
Recall & 0.93 & 0.64 & 0.83 & 0.76 & 0.89 & 0.92 & 0.79 \\
Precision & 0.77 & 0.60 & 0.73 & 0.63 & 0.56 & 0.73 & 0.28 \\
Bert & 0.78 & 0.40 & 0.74 & 0.59 & 0.42 & 0.79 & 0.33\\
\hline
\end{tabular}

\caption{AI-generated review multi-classification results(reported with F1 Score)}
\label{tab:LLM_self}
\end{table*}
}

%% file: sections/appendix.tex
\subsection{Examples for \AImodifiedimages }
\label{app:modify}

\begin{figure*}[h]
  \centering
  \includegraphics[width=0.9\textwidth, trim=0 80 0 0]{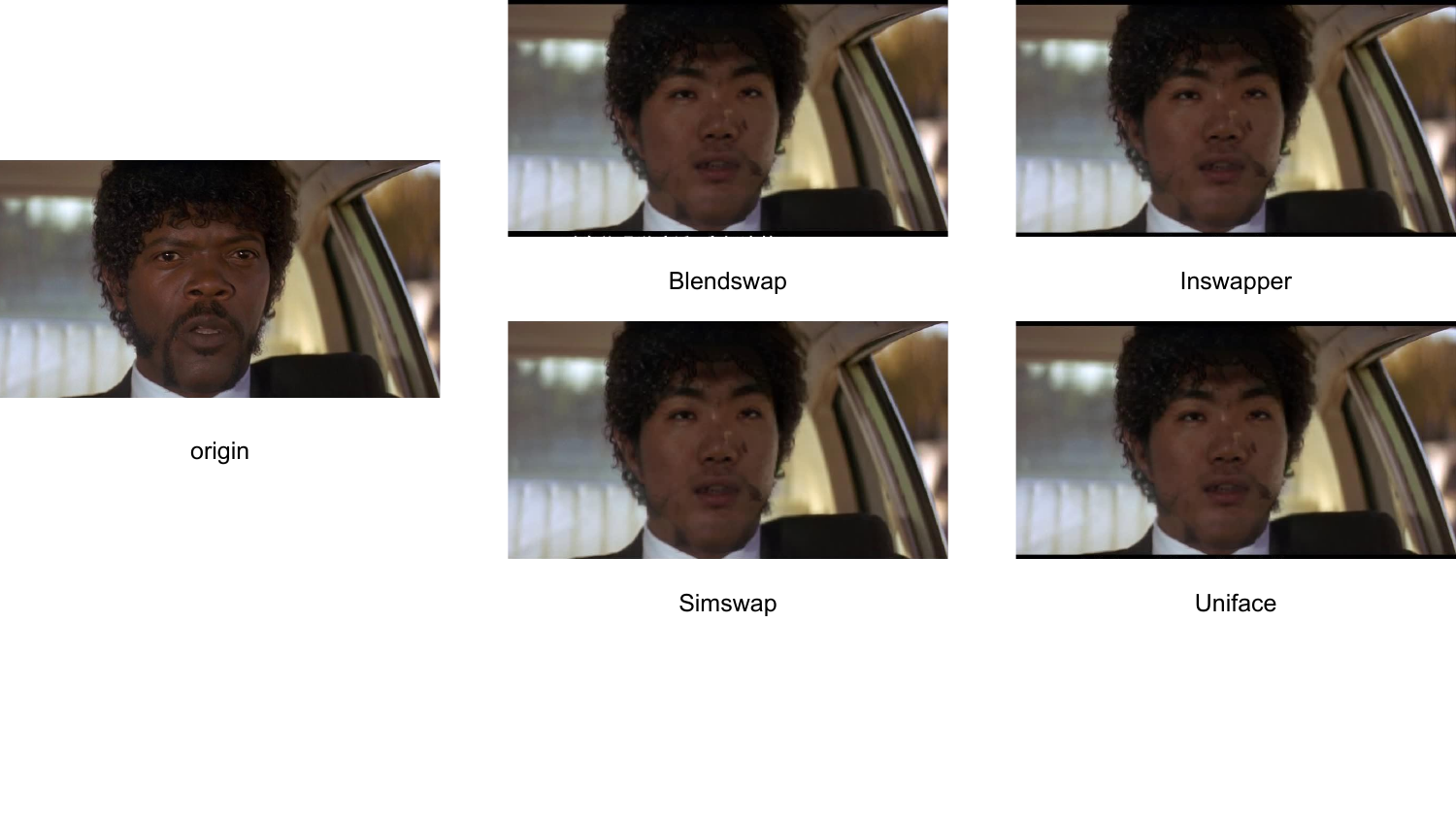}
  \caption{samples from \AImodifiedimages}
  \label{fig:swap}
\end{figure*}

\subsection{Prompts for dataset construction}
\label{app:dataconstruction}

\begin{table*}[ht]
\centering
\begin{tabular}{l}
\toprule
prompt\\
\midrule
Please generate a picture with the theme: \\

\{title\}\\\\

The picture should also following the description:\\

\{description\}\\

\bottomrule
\end{tabular}
\caption{prompts for generating fake images based on ImageNet captions.}
\label{tab:diff_prompt}
\end{table*}

\begin{table*}[h]
\centering
\begin{tabular}{l}
\toprule
prompt\\
\midrule
As an esteemed reviewer with expertise in the field of Artificial intelligence, you are asked to write \\a review for a scientific paper submitted for publication. Please follow the reviewer guidelines\\ provided below to ensure a  comprehensive and fair assessment: \\\\
In your review, you must cover the following aspects, adhering to the outlined guidelines:\\\\

Summary of the Paper: [Provide a concise summary of the paper, highlighting its main objectives,\\ methodology, results, and conclusions.]\\\\

Strengths and Weaknesses: [Critically analyze the strengths and weaknesses of the paper. \\Consider the significance of the research question, the robustness of the methodology, and the\\ relevance of the findings.]\\\\

Clarity, Quality, Novelty, and Reproducibility: [Evaluate the paper on its clarity of expression, \\overall quality of research, the novelty of the contributions, and the potential for reproducibility \\by other researchers.]\\\\

Summary of the Review: [Offer a brief summary of your evaluation, encapsulating \\your overall impression of the paper.]\\\\

Review Example 1:\\

\{human review\}\\

Follow the instructions above,and  write a review for the paper below:\\

\{paper\}\\

\bottomrule
\end{tabular}
\caption{prompts for generating fake reviews based on the papers.}
\label{tab:review_prompt}
\end{table*}
% Table \ref{tab:diff_prompt} presents the prompts designed for generating fake images based on ImageNet captions. 

% Table \ref{tab:review_prompt} presents the prompts designed for generating fake reviews based on the papers we collect.  The prompts provide a basic requirement for paper reviews and restrict the topics in each review. In addition, the prompts also give the explanation in for each topic and a random human review example.

\subsection{Distribution for image dataset }
\label{app:image}
\begin{minipage}{\textwidth}

\begin{minipage}[t]{0.5\textwidth}
\makeatletter\def\@captype{table}
\centering
\begin{tabular}{lc}
Models & Samples\\
\hline
Human & 10,000\\
Midjourney & 8,990 \\
Dall-e 3 & 9,649\\
Stable Diffusion & 9,996\\
Sdxl & 9,996\\
Flux & 9,996\\
\hline
overall & 58,627
\end{tabular}

\caption{Samples in AI-generated image dataset.}
\label{tab:AI_generated_image_dataset}

\end{minipage}
\begin{minipage}[t]{0.5\textwidth}
\makeatletter\def\@captype{table}
\centering
\begin{tabular}{lc}

Models & Samples\\

\hline
Human & 5,787\\
inswapper128 & 5,691\\
simswap256 &  5,691\\
uniface256 &  5,691\\
blendswap256 & 5,691\\

\hline
&\\
overall & 28,551
\end{tabular}

\caption{Samples in AI-modified image dataset.}
\label{tab:AI_modified_image_dataset}

\end{minipage}
\end{minipage}

\subsection{Raw review results}

\label{app:rawreview}
\begin{table}[!h]
\centering
\small
\begin{tabular}{lcccccccc}
\hline
Models & Human & GPT & Claude  & Gemini & DeepSeek & Llama3 & Ministral & overall\\
\hline
LR & 0.99 & 0.98 & 0.99 & 0.99 & 0.98 & 0.99 & 0.96 & 0.98\\
FCN & 0.98 & 0.97 & 0.99 & 0.97 & 0.96 & 0.98 & 0.98 & 0.97 \\
GPT2 & 0.99 & 0.99 & 0.99 & 0.99 & 0.99 & 0.98 & 0.98 & 0.98\\
Bert & 0.97 & 0.98 & 0.97 & 0.97 & 0.96 & 0.96 & 0.99 & 0.97\\
\hline
\end{tabular}
\caption{Performance of \AIpaperreview~with ``format-diverse'' (\textcolor{red}{malicious use})}
\label{tab:result:rawreview}
\end{table}

\subsection{Raw explain }

\label{app:explain}
\begin{table}[h]
    \centering
    % 调整行距（可选）
    \renewcommand{\arraystretch}{1.2}
    % 将第三列改为 p{8cm}，控制宽度并自动换行
    \begin{tabular}{c|c|p{10cm}}
    \hline
    \textbf{model} & \textbf{batch} & \textbf{explanation} \\
    \hline
    DALL-E 3 & 1 & \textbf{DALL‑E} tends toward whimsical or heightened-expression scenes and slightly simplified backgrounds.\\

    DALL-E 3 & 2 & \textbf{DALL‑E} Cheerful, cartoon-like image with big eyes, bright colors, and soft transitions.\\

    DALL-E 3 & 3 & \textbf{DALL‑E} tends toward idealized realism with smooth transitions and “picture-perfect” scenes.\\

    DALL-E 3 & 4 & \textbf{DALL‑E} Balanced realism with a “picturesque,” somewhat polished look; smooth transitions and bright but slightly airbrushed backgrounds.\\

    DALL-E 3 & 5 & \textbf{DALL‑E} Clean, near-photoreal scenes with a slightly “impossible” or whimsical twist.\\
    \hline
    Flux & 1 & \textbf{Flux} appears more straightforwardly photorealistic, with crisp details and little sign of AI distortion.\\

    Flux & 2 & \textbf{Flux} Semi-realistic shark scene with playful, anthropomorphic expression and clean lighting.\\

    Flux & 3 & \textbf{Flux} merges realistic details with playful stylization and vibrant colors.\\

    Flux & 4 & \textbf{Flux} Crisp, semi-realistic focus on the subject, vibrant colors, and a polished or “studio-lit” feel.\\

    Flux & 5 & \textbf{Flux} Crisp, realistic subjects with polished detail and refined focus, often using selective background blur.\\
    \hline
    Midjourney & 1 & \textbf{Midjourney} leans into cinematic, stylized, or dramatic aesthetics (the shark portrait).\\

    Midjourney & 2 & \textbf{Midjourney} Cinematic stingray illustration with dramatic light, textured detail, and painterly flair.\\

    Midjourney & 3 & \textbf{Midjourney} is known for high-impact, artistic compositions with rich detail and accentuated lighting.\\

    Midjourney & 4 & \textbf{Midjourney} Highly detailed, imaginative, and often fantasy-driven with rich texturing and painterly flair.\\

    Midjourney & 5 & \textbf{Midjourney} Bold, vibrant color usage and painterly aesthetics—strong fantasy or art-illustration feel.\\
    \hline
    sdxl & 1 & \textbf{SDXL} often produces high-resolution images with a painterly or illustrated quality and detailed layering of elements (the underwater shark).\\

    sdxl & 2 & \textbf{SDXL} A bold, high-resolution angry-face collage showcasing sharp outlines, graphic styling, and intense color saturation.\\

    sdxl & 3 & \textbf{SDXL} yields large-resolution images with a painterly or illustrated flair, and very sharp detail.\\

    sdxl & 4 & \textbf{SDXL} Capable of very sharp, stylized illustrations, including comic- or poster-like line art.\\

    sdxl & 5 & \textbf{SDXL} Very high-resolution detail and a distinct stylized or graphic approach (poster-like, thick outlines, bold color blocks).\\
    \hline
    Stable Diffusion & 1 & \textbf{Stable Diffusion} merges a decent level of realism with softer, diffused textures and repeating pattern artifacts (the floral backdrop in the woman’s portrait).\\

    Stable Diffusion & 2 & \textbf{Stable Diffusion} An anime-style portrait with a subdued palette, softly diffused brushstrokes, and stylized proportions.\\

    Stable Diffusion & 3 & \textbf{Stable Diffusion} typically delivers a more conventional photo-real aesthetic, though small artifacts or textural inconsistencies may appear upon close inspection.\\

    Stable Diffusion & 4 & \textbf{Stable Diffusion} Often yields more standard, photorealistic images, especially for nature shots, with subtle artifacts possible under close scrutiny.\\

    Stable Diffusion & 5 & \textbf{Stable Diffusion} Photographic realism in everyday contexts, often with subtle artifacts—frequently looking like a real camera shot.\\
    \hline
    \end{tabular}
    \caption{Description about different LMM outputs}
    \label{tab:image_desc}
\end{table}

\begin{table}[h]
    \centering
    \begin{tabular}{c|c|p{10cm}}
    \hline
    \textbf{model} & \textbf{batch} & \textbf{explanation} \\
    \hline
    gpt‑4o & 1 & \textbf{gpt‑4o} Well-structured headings with, carefully segmented sections for strengths and weaknesses, clarity and quality, and a concluding summary; methodical enumerations.\\
    gpt‑4o & 2 & \textbf{gpt‑4o} Formally sectioned: “Summary of the Paper,” “Strengths/Weaknesses,”.\\
    gpt‑4o & 3 & \textbf{gpt‑4o} Straight, methodical academic style.\\
    gpt‑4o & 4 & \textbf{gpt‑4o} Calm, advanced academic tone, methodical sectioning, ends with final summary or recommendation.\\
    gpt‑4o & 5 & \textbf{gpt‑4o} Often ends with a final summary or recommendation.\\
    \hline
    claude‑3 & 1 & \textbf{claude‑3} Tends to start with “Here’s my review…” or similarly direct intros, then a thorough set of bullet points, concluding with a summative statement.\\

    claude‑3 & 2 & \textbf{claude‑3} Opens with “Here’s my comprehensive review….”\\

    claude‑3 & 3 & \textbf{claude‑3} Maintains a balanced, conversational-yet-professional style.\\

    claude‑3 & 4 & \textbf{claude‑3} Clear subsections for Summary, Strengths, Weaknesses, plus a concluding recommendation or rating.\\

    claude‑3 & 5 & \textbf{claude‑3} Balanced bullet points, overall polite, constructive commentary, often includes a final numeric rating or acceptance.\\
    \hline
    gemini‑1.5‑pro & 1 & \textbf{gemini} Traditional academic headings; succinct bullet lists; balanced, formal tone.\\

    gemini‑1.5‑pro & 2 & \textbf{gemini} Organized with double-hash headings and short bullet points under “Strengths,” “Weaknesses.”\\

    gemini‑1.5‑pro & 3 & \textbf{gemini} Formal, academically tidy, ends with a recommendation. Less flamboyant than GPT-4, typically more concise.\\

    gemini‑1.5‑pro & 4 & \textbf{gemini} Bullet points for strengths and weaknesses are concise and direct.\\

    gemini‑1.5‑pro & 5 & \textbf{gemini} Ends with a reasoned acceptance recommendation and potential expansions for future work.\\
    \hline
    deepseek‑r1 & 1 & \textbf{deepseek} More essay-like paragraphs under each heading; uses bold headings but with lengthier narrative per bullet.\\

    deepseek‑r1 & 2 & \textbf{deepseek} More condensed, straightforward bullet lists focusing on “Key Contributions,” “Experiments,” “Practical Implications.”\\

    deepseek‑r1 & 3 & \textbf{deepseek} Minimal whimsical language, tends to zero in on method specifics and short paragraphs.\\

    deepseek‑r1 & 4 & \textbf{deepseek} Succinct bullet-like paragraphs focusing on method details and “Key Contributions,” “Limitations,” etc.\\

    deepseek‑r1 & 5 & \textbf{deepseek} More essay-like paragraphs under each heading; uses bold headings but with lengthier narrative per bullet.\\
    \hline
    llama‑3 & 1 & \textbf{llama‑3} Straight to the point, rarely uses decorative language, minimal headings, shorter paragraphs.\\

    llama‑3 & 2 & \textbf{llama‑3} Distinct comedic or thematic flair: “Arrr, ye landlubbers!” “scurvy dogs.”\\

    llama‑3 & 3 & \textbf{llama‑3} Mixes informal “pirate speak” with standard academic content. Possibly ends with whimsical remarks or comedic rating scale.\\

    llama‑3 & 4 & \textbf{llama‑3} Telltale informal or “pirate” language usage (“o’” in place of “of,” “Arrr…”).\\

    llama‑3 & 5 & \textbf{llama‑3} Balanced academic structure, but with a casual, often playful tone.\\
    \hline
    ministral & 1 & \textbf{ministral} A clear hierarchical heading structure, with a focus on enumerating strengths/weaknesses in short sub-sections; moderately formal.\\

    ministral & 2 & \textbf{ministral} Stream-of-consciousness style, with odd or tangential references.\\

    ministral & 3 & \textbf{ministral} Less coherent academic flow, more random jargon (“PyTorah,” “vanilla extensions,” “global horse-to-research-encoding network”).\\

    ministral & 4 & \textbf{ministral} Lacks the formal headings or bullet structures typical of the other models.\\

    ministral & 5 & \textbf{ministral} Often references domain knowledge or user logs in a broad, sometimes disjointed manner.\\
    \hline
    \end{tabular}
    \caption{Description about different LLM outputs}
    \label{tab:text_desc}
\end{table}